\documentclass[final]{cvpr}

\usepackage{times}
\usepackage{epsfig}
\usepackage{graphicx}
\usepackage{amsmath}
\usepackage{amssymb}
\usepackage{multirow}
\usepackage{tabularx}
\usepackage{makecell}
\usepackage{booktabs}
\usepackage{subfigure}
\usepackage{overpic}

\usepackage[pagebackref=true,breaklinks=true,colorlinks,bookmarks=false]{hyperref}

\def\cvprPaperID{1461} 
\def\confYear{CVPR 2021}

\begin{document}

\title{Bilateral Grid Learning for Stereo Matching Networks}

\author{Bin Xu$^{1}$, Yuhua Xu$^{1,2,}$\thanks{Corresponding author}~, Xiaoli Yang$^{1}$, Wei Jia$^{2}$, Yulan Guo$^{3}$
\\$^{1}$Orbbec Research,
$^{2}$Hefei University of Technology,
$^{3}$Sun Yat-sen University\\
 {\tt\small xyh\_nudt@163.com, guoyulan@sysu.edu.cn}
}

\maketitle
\thispagestyle{empty}
\pagestyle{empty}

\begin{abstract}
Real-time performance of stereo matching networks is important for many applications, such as automatic driving, robot navigation and augmented reality (AR). Although significant progress has been made in stereo matching networks  in recent years, it is still challenging to balance real-time performance and accuracy.
In this paper, we present a novel edge-preserving cost volume upsampling module based on the slicing operation in the learned bilateral grid. The slicing layer is parameter-free, which allows us to obtain a high quality cost volume of high resolution from a low-resolution cost volume under the guide of the learned guidance map efficiently. The proposed cost volume upsampling module can be seamlessly embedded into many existing stereo matching networks, such as GCNet, PSMNet, and GANet. The resulting networks are accelerated several times while maintaining comparable accuracy. Furthermore, we design a real-time network (named BGNet) based on this module, which outperforms existing published real-time deep stereo matching networks, as well as some complex networks on the KITTI stereo datasets. The code is available at \textcolor{magenta}{https://github.com/YuhuaXu/BGNet}.
\end{abstract}

\section{Introduction}\label{sec:intro}

Stereo matching is a key step in 3D reconstruction, which has numerous applications in the fields of 3D modeling, robotics, UAVs, augmented realities (AR), and autonomous driving~\cite{scharstein2014high, kitti2012, bao2020instereo2k}. 
Given a pair of stereo images, the purpose of stereo matching is to establish dense correspondences between the pixels in the left and right images. 
Although this problem has been studied for more than 40 years, it has not been completely solved due to some difficult factors, such as sensor noise, foreground-background occlusion, weak or repeated textures, reflective regions, and transparent objects. 

In recent years, deep learning has shown great potential in this field~\cite{mc_cnnn2016stereo, gcnet2017, iresnet2018stereo, stereonet2018, PSMNet2018, AANet2020}. Some recent work indicates that 3D convolutions can improve the accuracy of the disparity estimation networks~\cite{gcnet2017, PSMNet2018, deeppruner2019}. However, the 3D convolutions are time consuming, which limits their application in real-time systems.

Bilateral filter~\cite{bilateral1998} is an edge-preserving filter, which has wide applications in image de-noising, disparity estimation~\cite{asw2006adaptive}, and depth upsampling~\cite{yang2013fusion}. 
However, the original implementation of bilateral filter is time-consuming. 
Bilateral grid proposed by Chen \etal~\cite{bilateral_grid2007} is a technique for speeding up bilateral filtering. It treats the filter as a ``splat/blur/slice" procedure. That is, pixel values are ``splatted" onto a small set of vertices in a grid, those vertex values are then blurred, and finally the filtered pixel values are produced via a ``slice" (an interpolation) of the blurred vertex values. 
Recently, Gharbi \etal ~\cite{hdrnet2017} utilize the idea of the bilateral grid to estimate the local affine color transforms in their real-time image enhancement network. 

For StereoNet \cite{stereonet2018}, the 2D disparity map regressed from an aggregated 4D cost volume at a low resolution (e.g., 1/8) is upsampled via bilinear interpolation and refined hierarchically. Although the speed of the lightweight network is fast, the accuracy is relatively low compared with existing complex networks, as shown in Table \ref{tab:kitti12012_2015}. On the ranking list of KITTI 2012 \cite{kitti2012} and KITTI 2015 \cite{kitti2015}, the top performing methods usually conduct 3D convolutions at a relatively high resolution, such as 1/3 resolution for GANet \cite{ganet2019} and 1/4 resolution for PSMNet \cite{ganet2019}. However, their efficiency is reduced. For example, it takes 1.8s for GANet to process a pair of images of 1242 $\times$ 375, and 0.41s for PSMNet. 

The motivation of this paper is to propose a solution that can regress the disparity map at a high resolution to keep the high accuracy, while maintaining high efficiency. 
The main contributions of this work are as follows:

(1) We design a novel edge-preserving cost volume upsampling module based on the learned bilateral grid. It can efficiently obtain a high-resolution cost volume for disparity estimation from a low-resolution cost volume via a slicing operation. 
With this module, cost volume aggregations (e.g., 3D convolutions) can be performed at a low resolution. The proposed cost volume upsampling module can be seamlessly embedded into many existing networks such as GCNet, PSMNet, and GANet. The resulting networks can be accelerated 4$\sim$29 times while maintaining comparable accuracy. To the best of our knowledge, it is the first time that the differential bilateral grid operation is applied in deep stereo matching networks.

(2) Based on the advantages of the proposed cost volume upsampling module, we design a real-time stereo matching network, named BGNet, which can process stereo pairs on the KITTI 2012 and KITTI 2015 datasets at 39fps. Experimental results show that BGNet outperforms existing published real-time deep stereo matching networks, as well as some complex networks, such as GCNet, AANet, DeepPruner-Fast, and FADNet, on the KITTI 2012 and KITTI 2015 stereo datasets.


\section{Related Work}
\textbf{Stereo Matching Network.} MC-CNN~\cite{zbontar2015computing} is the first work that uses convolutional neural network (CNN) to compare two image patches (e.g., 11$\times$11) and calculate their matching costs. Meanwhile, the following steps, such as cost aggregation, disparity computation, and disparity refinement, are still traditional methods~\cite{adcensus2011building}. 
It significantly improves the accuracy, but still struggles to produce accurate disparity results in textureless, reflective and occluded regions and is time-consuming. 
DispNetC~\cite{dispNetC2016large} is the first end-to-end stereo matching network with a similar network structure as FlowNet~\cite{flownet2015}. DispNetC is more efficient, almost 1000 times faster than MC-CNN-Acrt~\cite{zbontar2015computing}. In DispNetC, there is an explicit correlation layer. To further improve the estimation accuracy, the residual refinement layers are exploited \cite{iresnet2018stereo, liang2019stereo, cascade2017}. Besides, the segmentation information \cite{segstereo2018} and the edge information \cite{edgestereo2020} are incorporated into the networks to improve the performance.
GC-Net~\cite{kendall2017end} uses 3D convolutions for cost aggregation in a 4D cost volume, and utilizes the soft $argmin$ to regress the disparity. 
DeepPruner \cite{deeppruner2019} brings the idea of PatchMatch Stereo \cite{patchmatch2011}, and builds a narrow cost volume based on the estimated lower and upper bounds of the disparity to speed up the prediction. The narrow cost volume optimization is also used in \cite{cascade_cost_volume2020}. 
Since disparities can vary significantly for stereo cameras with different baselines, focal lengths and resolutions, the fixed maximum disparity used in cost volume hinders them to handle different stereo image pairs with large disparity variations. Wang \etal \cite{wang2019learning,wang2020parallax} propose a generic parallax-attention mechanism to capture stereo correspondence regardless of disparity variations.

Recent work~\cite{PSMNet2018, deeppruner2019} shows that the networks with 3D convolution can achieve higher disparity estimation accuracy on specific datasets. However, 3D convolution is more time-consuming than 2D convolution, which makes it difficult to apply in real-time applications. In order to pursue real-time performance, StereoNet~\cite{stereonet2018} performs 3D convolution at a low resolution (e.g., 1/8 resolution) and the resulting network can run in real-time at 60fps. However, this simplification reduces the network's accuracy.
For other networks, such as AANet~\cite{AANet2020}, FADNet~\cite{fadnet2020}, and DeepPruner-Fast~\cite{deeppruner2019}, although the accuracy has been improved, they have not achieved real-time performance.

\textbf{Bilateral Grid.} Since this work is inspired by bilateral grid of Chen \etal\cite{bilateral_grid2007}, we will give it a brief review. 
\par
The bilateral grid was originally introduced to speed up the bilateral filter. It consists of three steps, including splatting, blurring, and slicing. 
For an input image $I$ and a guidance image $G$, the splatting operation projects the original pixels of $I$ into a 3D grid $\mathcal{B}$, where the first two dimensions $(x, y)$ correspond to 2D position in the image plane and form the spatial domain, while the third dimension $g$ corresponds to the image intensity of the guidance image. Then,  Gaussian blurring is performed in the 3D bilateral grid $\mathcal{B}$. 
Finally, based on the blurred bilateral grid and the guidance image $G$, a 2D value map $\bar{I}$ is extracted by accessing the grid at $(sx, sy, s_GG(x, y))$ using trilinear interpolation, where
$s$ is the width or height ratio of the grid's dimension w.r.t the original image dimension, and $s_G\in(0,1)$ is the ratio of the gray level of the grid to the gray level of the guidance image $G$. 
This linear interpolation under the guide of the guidance image in the bilateral grid is called slicing. In practice, most operations on the grid require only a coarse resolution, where the number of grid cells is much smaller than the number of image pixels. See \cite{bilateral_grid2007} for more details.

The bilateral grid is utilized to accelerate stereo matching algorithms~\cite{richardt2010real, barron2015fast}.
Chen \etal \cite{chen2016bilateral} approximate an image operator with a grid of local affine models in bilateral space, the parameters of which are fit to a reference input and output pair. By performing this model-fitting on a low-resolution image pair, this technique enables real-time on-device computation.
Gharbi \etal \cite{hdrnet2017} build upon this bilateral space representation. Rather than fitting a model to approximate a single instance of an operator from a pair of images, they construct a rich CNN-like model that is trained to apply the operator to any unseen input.

\section{Method}

Inspired by bilateral grid processing~\cite{bilateral_grid2007}, we propose an edge-aware cost volume upsampling module. 
With this module, we can perform the majority of calculation at a low resolution. Meanwhile, we can obtain accurate disparity prediction with the upsampled cost volume at a high resolution. 
In this section, we first describe the proposed cost volume upsampling module in details. Then, we show that the upsampling module can be utilized as an embedded module in many existing stereo matching networks. Additionally, based on the advantages of this module, we design a real-time stereo matching network.

\begin{figure}
\begin{center}
\includegraphics[width=1.0\linewidth]{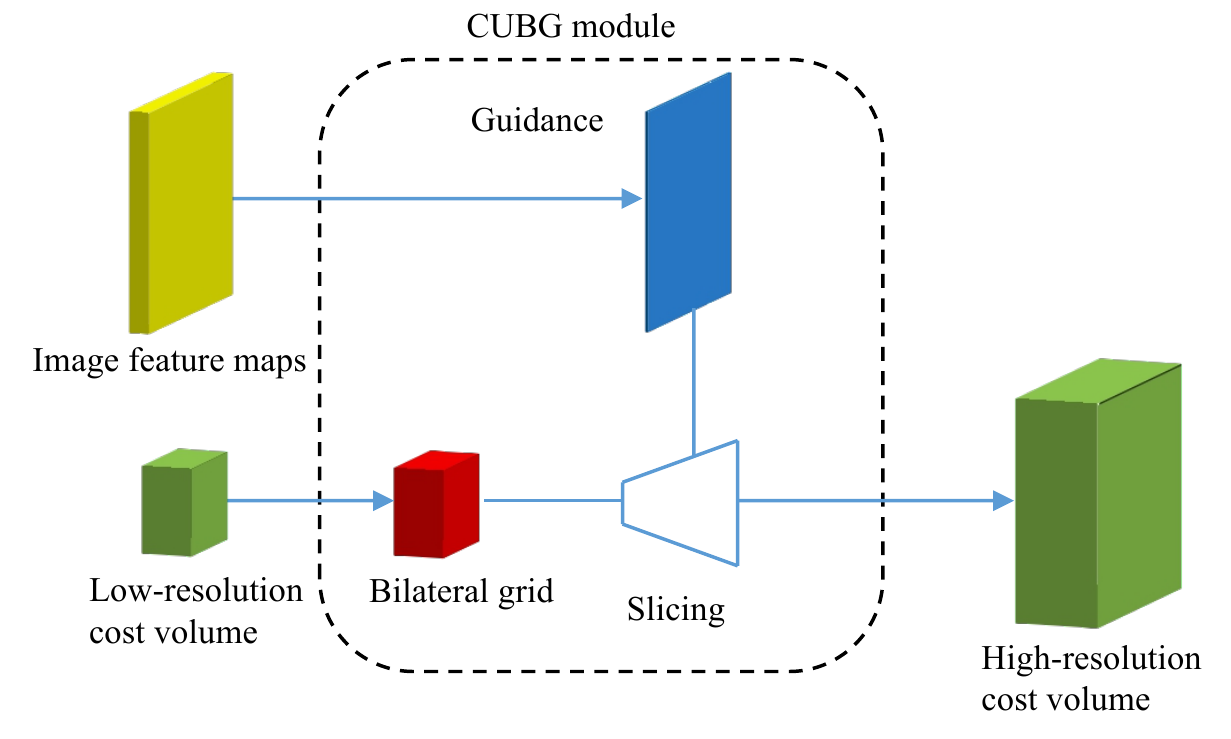}
\end{center}
\caption{The module of Cost volume Upsampling in the learned Bilateral Grid (CUBG). The high-quality cost volume at high resolution can be obtained with the CUBG module from the low resolution (e.g., 1/8) via the slicing operation of bilateral grid processing.}
\label{fig:bg_module}
\end{figure}

\begin{figure*}
\begin{center}
\includegraphics[width=1.0\linewidth]{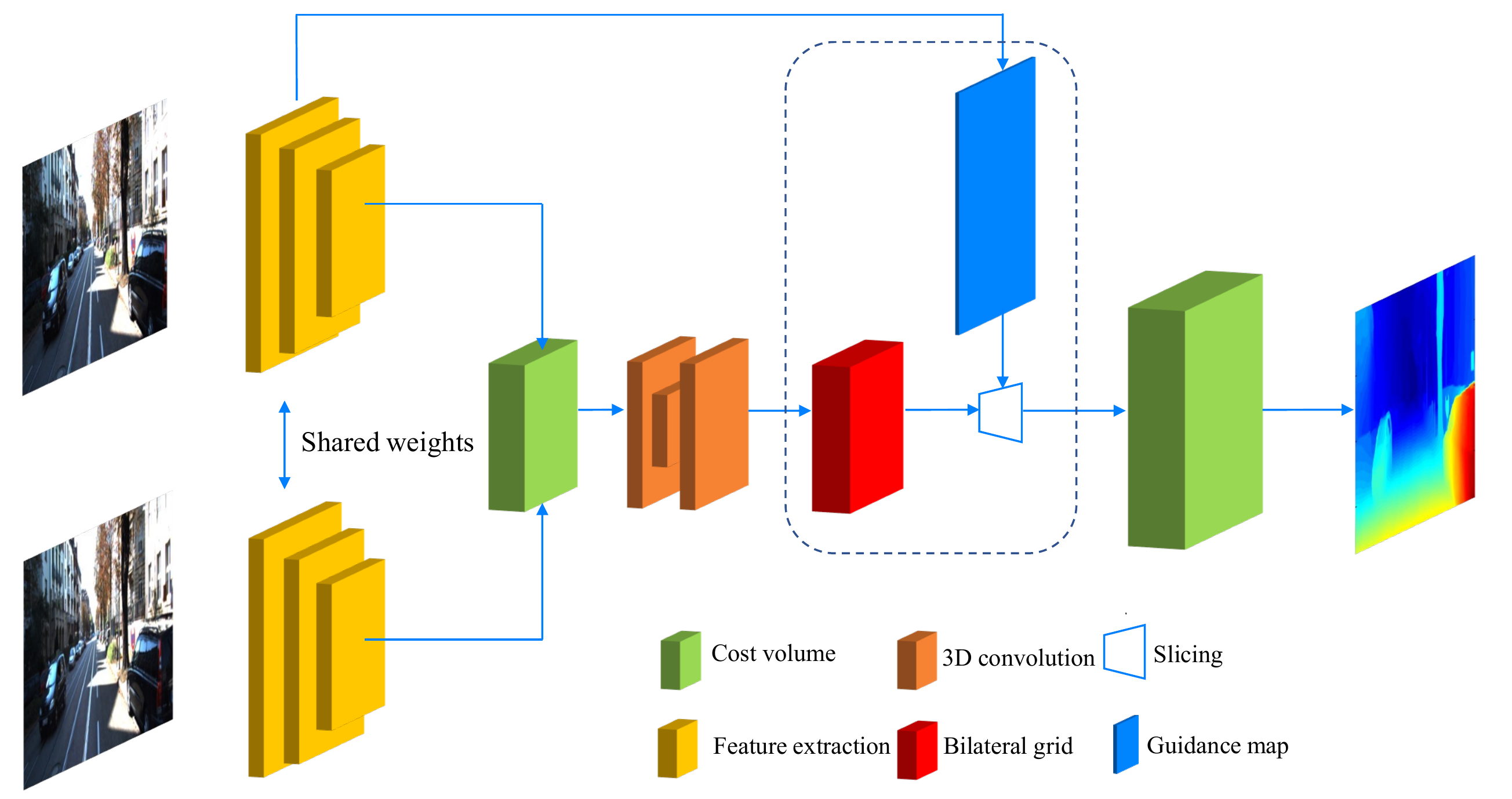}
\end{center}
\caption{Overview of the proposed BGNet. } 
\label{fig:bgnet}
\end{figure*}

\subsection{Cost Volume Upsampling in Bilateral Grid}

As stated in Section \ref{sec:intro}, the goal of this paper is to seek a solution that can regress the disparity map at a high resolution to keep the high accuracy, while maintaining high efficiency. 
In order to achieve this goal, we design a module of Cost volume Upsampling in Bilateral Grid (CUBG). The CUBG module can upsample the cost volume calculated at the low resolution (e.g., 1/8) to a high resolution via the slicing operation of bilateral grid processing.

As illustrated in Figure \ref{fig:bg_module}, the input of the CUBG module is a low-resolution cost volume $\mathcal{C_L}$ and image feature maps, and the output is the upsampled high-resolution cost volume. The operation in the CUBG module includes bilateral grid creation and slicing. 

\textbf{Cost volume as a bilateral grid.} 
Given an aggregated cost volume $\mathcal{C_L}$ with four dimensions (including width $x$, height $y$, disparity $d$ , and channel $c$) at a low resolution (e.g., 1/8), the conversion from the cost volume $\mathcal{C_L}$ to the bilateral grid $\mathcal{B}$ is straightforward.
In all experiments, we use a 3D convolution of 3 $\times$3$\times$3 to achieve the conversion. 

There are four dimensions in the bilateral grid $\mathcal{B}$: width $x$, height $y$, disparity $d$, and guidance feature $g$. The value in the bilateral grid is represented by $\mathcal{B}(x,y,d,g)$. 

\textbf{Upsampling with a slicing layer.}
With the bilateral grid, we can produce a 3D high-resolution cost volume $\mathcal{C}_H$ ($\mathcal{C}_H$ $\in$ $\mathbb{R}^{W,H,D}$) via a slicing layer. This layer performs data-dependent lookups in the low-resolution grid of matching costs. Specifically, the slicing operation is the linear interpolation in the four-dimensional bilateral grid under the guide of the 2D guidance map $G$ of high resolution. The slicing layer is parameter-free and can be implemented efficiently. Formally, the slicing operation is defined as

\begin{equation}
\mathcal{C}_H(x,y,d)=\mathcal{B}\left(sx,sy,sd,s_GG(x,y)\right)
\end{equation}
where $s\in(0,1)$ is the width or height ratio of the grid's dimension w.r.t the high-resolution cost volume dimension, $s_G\in(0,1)$ is the ratio of the gray level of the grid ($l_{grid}$) to the gray level of the guidance map $l_{guid}$. 

The guidance map $G$ is generated from the high-resolution feature maps via two $1 \times 1$ convolutions. Note that, the guidance information of each pixel depends on its own feature vector.
Consequently, sharp edges can be obtained.

\par
Unlike the original grid designed in \cite{bilateral_grid2007}, the bilateral grid in this work is learned from the cost volume automatically. In experiments, the size of the grid is usually set to $W/8\times H/8 \times D_{max}/8 \times 32$, where $W$ and $H$ are image width and height, respectively. $D_{max}$ is the maximal disparity value.

\subsection{Network Architecture}
\subsubsection{Embedded Module}\label{sec:embedded}
The proposed CUBG module can be seamlessly embedded into many existing end-to-end network architectures. 
In this work, we embed the CUBG module into four representative stereo matching models: GCNet \cite{gcnet2017}, PSMNet \cite{PSMNet2018}, GANet \cite{ganet2019}, and DeepPrunerFast \cite{deeppruner2019}. The resulting models are denoted with a suffix BG (e.g., GCNet-BG). For the first three networks, we re-build the cost volume at the resolution of 1/8, 1/8, and 1/6, respectively. Then, the cost volumes are upsampled via our CUBG module to the resolution of 1/2, 1/4, and 1/3, respectively. 
For DeepPrunerFast, the PatchMatch-like upper and lower bounds estimation module and the narrow cost volume aggregation module are replaced by a full cost volume aggregation at the resolution of 1/8. Then, the cost volume is upsampled via the CUBG module to the resolution of 1/2. All the other parts remain the same as in their original implementation.

\subsubsection{BGNet}\label{sec:bgnet}
Based on the CUBG module, we also design an efficient end-to-end stereo matching network, named BGNet.
Figure \ref{fig:bgnet} provides an overview of the network. 
The network consists of four modules of feature extraction, cost volume aggregation, cost volume upsampling, and residual disparity refinement. It can run in real-time at the resolution of the KITTI stereo dataset. In the following, we introduce these modules in details.

\textbf{Feature Extraction.} 
The ResNet-like architecture is widely used in stereo matching networks \cite{PSMNet2018, group_wise2019}. We utilize the similar architecture to extract the image features for matching here.
For the first three layers, three convolution of 3 $\times$ 3 kernel with strides of 2, 1, and 1 are used to downsample the input images. Then, four residual layers with strides of 1, 2, 2, and 1 are followed to quickly produce unary features at 1/8 resolution. Two hourglass networks are followed to obtain large receptive fields and rich semantic information. Finally, all the feature maps at 1/8 resolution are concatenated to form feature maps with 352 channels for the generation of the cost volume. We use $\mathbf{f}_l$ and $\mathbf{f}_r$ to represent the final feature maps extracted from the left and right images, respectively.

\textbf{Cost Aggregation.} After the feature extraction modules, we build a group-wise correlation cost volume \cite{group_wise2019} for aggregation, which combines the advantages of the concatenation volume and the correlation volume. 
Group-wise correlation is computed for each pixel location $(x,y)$ at disparity level $d$ by dividing the feature channels into $N_{\bar{g}}$ groups, the same as in \cite{group_wise2019}. In our experiments, $N_{\bar{g}}=44$.

In PSMNet \cite{PSMNet2018}, a stacked hourglass architecture was utilized to optimize the cost volume using the contextual information. Here, considering the efficiency,  only one hourglass architecture is used to filter the cost volume. Specifically, we first use two 3D convolutions to reduce the channel number of cost volume from 44 to 16. Then, a U-Net-like 3D convolution network is used for cost aggregation, where  skip connection is replaced by an element-wise summation operation to reduce computational cost. 
We use $\mathcal{C_L}$ to represent the aggregated cost volume at 1/8 resolution.

 \par
\textbf{Disparity Regression.} 
With the high-resolution cost volume $\mathcal{C}_H$, we can regress the disparity prediction via $soft$ $argmin$ as in \cite{gcnet2017}:
\begin{equation}
\mathbf{D}_{pred}(x,y)=\sum \limits_{d=0}^{D_{max}} d \times softmax\left(\mathcal{C}_H(x,y,d)\right)
\end{equation}


\par
\textbf{Loss Function.}
The loss function $L$ is defined on the final prediction $\mathbf{D}_{pred}$ using the smooth $L_1$  loss $\mathcal{L}$,
\begin{equation}
\begin{aligned}
L = \sum \limits_{p}\mathcal{L}\left(\mathbf{D}_{pred}(p)-\mathbf{D}_{gt}(p)\right)
\end{aligned}
\end{equation}
Here,
$$
\mathcal{L}(x)=\left\{\begin{aligned}&0.5x^2, &
{\rm if}~|x|<1\\
&|x|-0.5, &{\rm otherwise}\end{aligned}\right.
$$
where $\mathbf{D}_{gt}(p)$ is the ground truth disparity for pixel $p$.





\begin{figure}
\begin{center}
\includegraphics[width=1.0\linewidth]{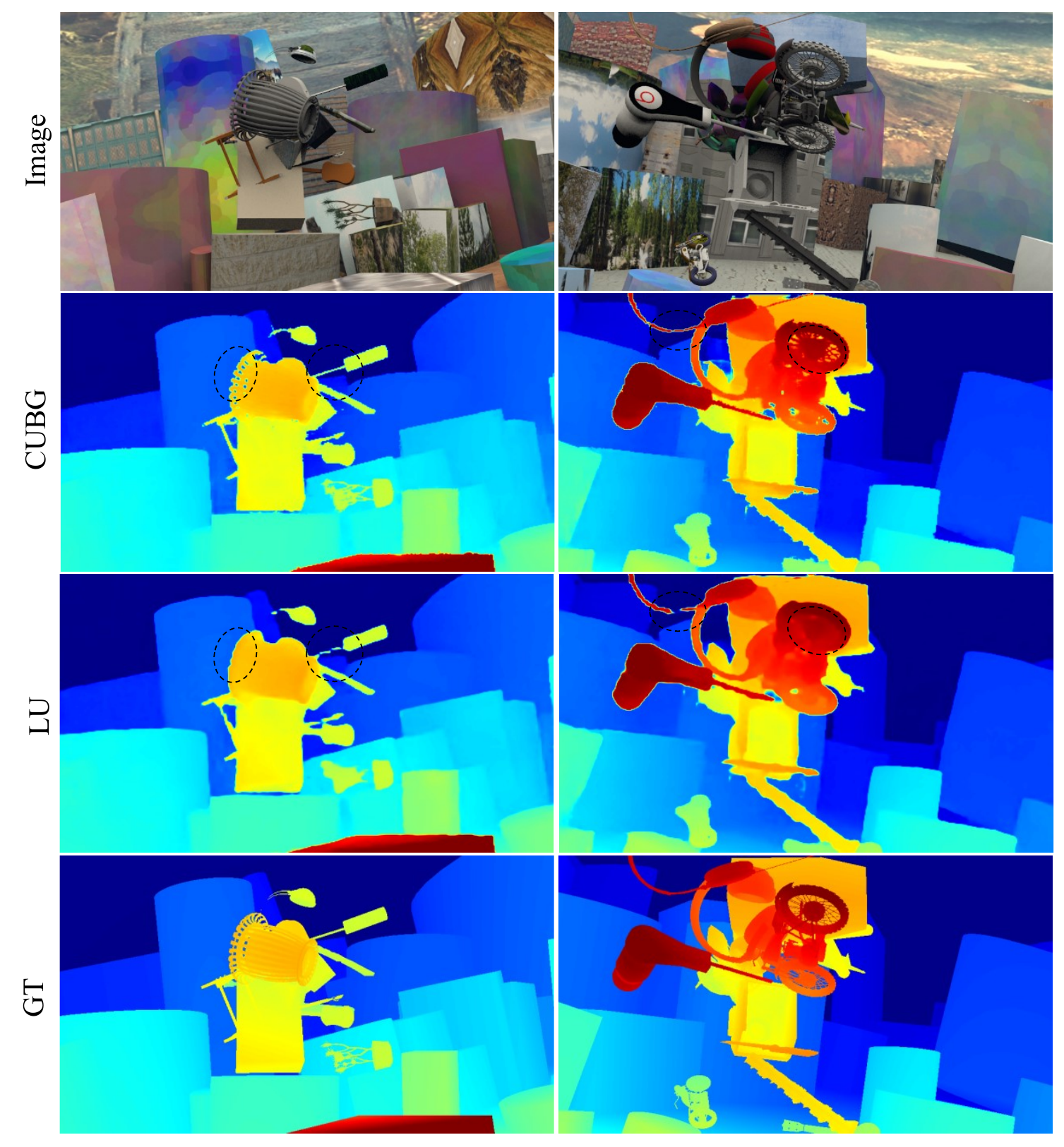}
\end{center}
\caption{Qualitative results on SceneFlow. The abbreviations Linear Upsampling (LU) and CUBG denote the two different cost volume upsampling methods.}
\label{fig:sceneflow}
\end{figure}

\section{Experimental Results}\label{sec:experiment}

\subsection{Datasets and Evaluation Metrics}
\par
\textbf{SceneFlow.} The synthetic stereo datasets include Flyingthings3D, Driving, and Monkaa. These datasets consist of 35,454 training images and 4,370 testing images of size 960$\times$540 with accurate ground-truth disparity maps. Since the Finalpass of the SceneFlow datasets contains more motion blur and defocus and is more like real-world images than the Cleanpass, we use the Finalpass for ablation study. The End-Point-Error (EPE) will be used as the evaluation metric for the SceneFlow dataset.

\textbf{KITTI.} KITTI 2012~\cite{kitti2012} and KITTI 2015~\cite{kitti2015} are outdoor driving scene datasets. KITTI 2012 provides 194 training and 195 testing image pairs, KITTI 2015 provides 200 training and 200 testing image pairs. The resolution of KITTI 2015 is 1242 $\times$ 375, and that of KITTI 2012 is 1226 $\times$ 370. The groundtruth disparities are obtained from LiDAR points.
For KITTI 2012, percentages of erroneous pixels and average end-point errors for both nonoccluded (\textit{Noc}) and all (\textit{All}) pixels are reported. 
For KITTI 2015, the percentage of disparity outliers \textit{D1} is evaluated. 

\textbf{Middlebury 2014.} The Middlebury stereo dataset \cite{middlebury2014} had been widely used for the evaluation of stereo matching methods.
The disparity maps in this dataset are calculated using accurate structured-light techniques. However, this dataset only contains dozens of image pairs and is insufficient to train a deep neural network.

We follow previous end-to-end approaches by performing initial training on SceneFlow and then individually fine-tuning the resulting model on the KITTI datasets.
For Middlebury, we only use the dataset for the evaluation of generalization ability without fine-tuning.

\subsection{Implementation Details}
Our BGNet is implemented with PyTorch on NVIDIA RTX 2080Ti GPU. We use the Adam optimizer with $\beta_{1} = 0.9$ and $\beta_{2} = 0.999$. 

For the SceneFlow dataset, a series of data augmentation techniques, including asymmetric chromatic augmentation, y-disparity augmentation \cite{hsm2019hierarchical}, Gaussian blur enhancement, and scale zoom enhancement are adopted. Each enhancement module has a 50\% chance of being used. In addition, following CRL \cite{crl2017cascade}, the stereo pairs with more than 25\% of their disparity values larger than 300 are removed. 
We use the one-cycle scheduler \cite{smith2019super} to adjust the learning rate with the maximum value of 0.001. The batch size is set to 16, and crop size is set to 512 $\times$ 256  to  train our network for 50 epochs in total. 
To evaluate these methods on the test dataset of SceneFlow, we set the maximum disparity values to 192. Pixels with disparity values out of the valid range are not considered in the evaluation.

For KITTI 2015, we randomly select 20 pairs as the validation set, and use the remaining 180 pairs and 194 pairs of KITTI 2012 as the training set. We first finetune our network with a constant learning rate of 0.001 for 300 epochs on the model pre-trained on SceneFlow, and repeat three times to pick the one with the best evaluation metrics.
For KITTI 2012, we use a similar finetuning strategy.


\subsection{Ablation Study}\label{Ablation_Study}
To validate the effectiveness of the proposed CUBG module, we first replace the module in BGNet with  direct linear upsampling on the cost volume, and evaluate the EPE results on SceneFlow. As shown in Table \ref{tab:epe_region}, EPE rises from \textbf{1.17} to \textbf{1.40}. 
Qualitative results on SceneFlow are shown in Figure \ref{fig:sceneflow}. Although the disparity map is reconstructed from a low-resolution cost volume, many thin structures and sharp edges are still recovered by BGNet, 
which benefits from the edge-aware nature of the bilateral grid. Performing disparity regression in a high-resolution cost volume obtained via a slicing layer in the bilateral grid forces the predictions of our network to follow the edges in the guidance map $G$, thereby regularizing our predictions towards edge-aware solutions.

To validate this point, we also evaluate the methods in flat regions and the regions near edges.
Edges are first detected with the Canny detector \cite{canny1986} in ground-truth disparity maps. Then, these detected edges are dilated with a $5\times5$ square structuring element. The EPE of the regions of these dilated edges is denoted as EPE-edge, others are denoted as EPE-flat. 
Quantitative results are shown in Table \ref{tab:epe_region}, from which we have two major observations. 
First, for both methods, EPE-edge is much higher than EPE-flat, which means the disparities near edges are hard to estimate.
Second, in flat regions, the errors of these two methods are comparable. However, in the regions near edges, the EPE of CUBG is 2.18 lower than linear upsampling, which benefits from the edge-aware property of CUBG. 
Figure \ref{fig:sceneflow} shows qualitative comparison.  When the CUBG module is utilized, the predicted edges are more accurate and sharper  than linear upsampling, and the thin structures are better reconstructed.

We also performed ablation study on Middlebury 2014 and KITTI 2015. For Middlebury 2014, we use 13 additional datasets with GT for training (finetuning), where 78 pairs of images are available in total. 
For the KITTI 2015 dataset, where the training set is split into 160 pairs for training and 40 pairs for validation (as done in AANet~\cite{AANet2020}).
For Middlebury 2014, the Bad 2.0 errors for LU and BG are 18.7\% and 16.8\% (48.9\% vs 45.3\% near edges, and 14.7\% vs 13.0\% in flat regions).
For KITTI 2015, the D1-all errors for LU and BG are 2.14\% and 2.01\%, respectively.

{\bf Impact of guidance map.} If the original guidance map is replaced with the luma-version of the input image, the EPE on Scene Flow increases from 1.17 to 1.28. 

{\bf Building cost volume at different resolutions.} When building the cost volume at the resolution of 1/16, the EPE on Scene Flow increases from 1.17 to 1.58, and the run-time decreases from  25.3 ms to 17.1 ms. 

\begin{table} 
\begin{center}
\begin{tabular}{|c|c|c|c|c|}
\hline
Method & EPE & EPE-edge & EPE-flat &Time (ms)\\ 
\hline
CUBG & 1.17 &  5.95 & 0.68 & 25.3 \\
LU & 1.40 & 8.13 & 0.71 & 25.1\\
\hline
\end{tabular}
\end{center}
\caption{Ablation study results of the proposed networks on Finalpass of the SceneFlow datasets \cite{dispNetC2016large}. The abbreviation LU denotes linear upsampling. EPE-edge represents the EPE near edges of objects, and EPE-flat represents the EPE in flat regions.}\label{tab:epe_region}
\end{table}

\textbf{Embedded module.} Table \ref{tab:embedded} shows the quantitative results of GCNet-BG, PSMNet-BG, GANet-BG, and DeepPruner-BG, where the CUBG module is used as an embedded module in these networks.  
Compared with top-performing stereo models GC-Net, PSMNet and GA-Net, our method not only obtains clear performance improvements, but is also significantly faster (29$\times$
than GC-Net, 4.9$\times$ than PSMNet, and 4.2$\times$ than GANet), demonstrating the high efficiency of the CUBG module. Additionally, although the cost volumes of  DeepPrunerFast and DeepPrunerFast-BG  have the same resolution, DeepPrunerFast-BG is faster while having lower EPE.


\begin{table} 
\begin{center}
\begin{tabular}{|c|c|c|c|c}
\hline
Method & Res-CV & EPE &  \makecell{Time \\(ms)}\\ 
\hline
GCNet~\cite{gcnet2017} &    1/2 & 2.51    & 1673.2 \\
GCNet-BG & 1/8 & \textbf{1.07} & \textbf{57.1} \\
\hline
PSMNet~\cite{PSMNet2018}    & 1/4 & 1.09   & 439.6\\
PSMNet-BG & 1/8 & \textbf{0.92}  & \textbf{89.4}\\
\hline
GANet\_deep~\cite{ganet2019}    & 1/3 & 0.95   & 2240.2 \\
GANet\_deep-BG & 1/6 & \textbf{0.63}  & \textbf{533.4} \\
\hline
DeepPrunerFast~\cite{deeppruner2019}    & 1/8 & 0.97  & 64.7 \\
DeepPrunerFast-BG & 1/8 & \textbf{0.84}  & \textbf{56.6} \\
\hline
\end{tabular}
\end{center}
\caption{Evaluation of the CUBG module embedded into other networks on the SceneFlow dataset \cite{dispNetC2016large}. We embed the CUBG module into four representative stereo matching models, GCNet, PSMNet, GANet, and DeepPrunerFast. Resulting models are denoted with a suffix BG. Res-CV represents the resolution at which the cost volume is built.}
\label{tab:embedded}
\end{table}


We also tested them on the KITTI 2015 and Middlebury 2014 datasets, where the finetuning strategies are the same as those in Subsection \ref{Ablation_Study}.
The results in Table \ref{tab:embedded-kitti2015} show that, compared with their original networks, GA-Net-BG and PSM-Net-BG achieve comparable accuracy while being accelerated more than four times.
 

\begin{table} 
\begin{center}
\begin{tabular}{|c|c|c|c|c|c|}
\hline
Method & Res & \makecell{D1-all\\(KIT)} & \makecell{T (ms) \\(KIT)} & \makecell{Bad2.0\\(MID)}\\
\hline
PSMNet \cite{PSMNet2018}   & 1/4 & \textbf{1.95} & 410    & \textbf{19.36}\\
PSMNet-BG & 1/8 & 2.07& \textbf{79}   & 20.77\\ 
\hline
GANet\_deep  \cite{ganet2019}  & 1/3 & \textbf{1.58} & 1800   & 15.67\\
GANet\_deep-BG & 1/6 & 1.67 & \textbf{406}   & \textbf{15.30} \\
\hline
DeepPrunerFast \cite{deeppruner2019}   & 1/8 & 2.06 & 61  & 16.80 \\
DeepPrunerFast-BG & 1/8 & \textbf{1.91} & \textbf{55}  & \textbf{15.14}\\
\hline
\end{tabular}
\end{center}
\caption{Evaluation of the CUBG module when embedded into three networks on the KITTI 2015 and Middlebury 2014 datasets. The second column gives the resolution at which the cost volume is built, the third and fourth columns show D1-all and run-time results on KITTI 2015 (KIT), and the last column shows Bad 2.0 errors on Middlebury (MID).}
\label{tab:embedded-kitti2015} 
\end{table}

\subsection{Evaluation on KITTI}
Table \ref{tab:kitti12012_2015} shows the performance and runtime of competing algorithms on the KITTI stereo benchmark. 
Our BGNet can run in real-time on the dataset at \textbf{39} fps. 
Compared with the latest FADNet~\cite{fadnet2020}, AANet~\cite{AANet2020} and DeepPruner-Fast~\cite{deeppruner2019}, our BGNet is not only more accurate, but also two times faster than those methods, as shown in Table \ref{tab:kitti12012_2015}.
Among the published networks with computational time less than 50 ms, our BGNet has the best accuracy. 

\begin{figure}
\begin{center}
\includegraphics[width=1.0\linewidth]{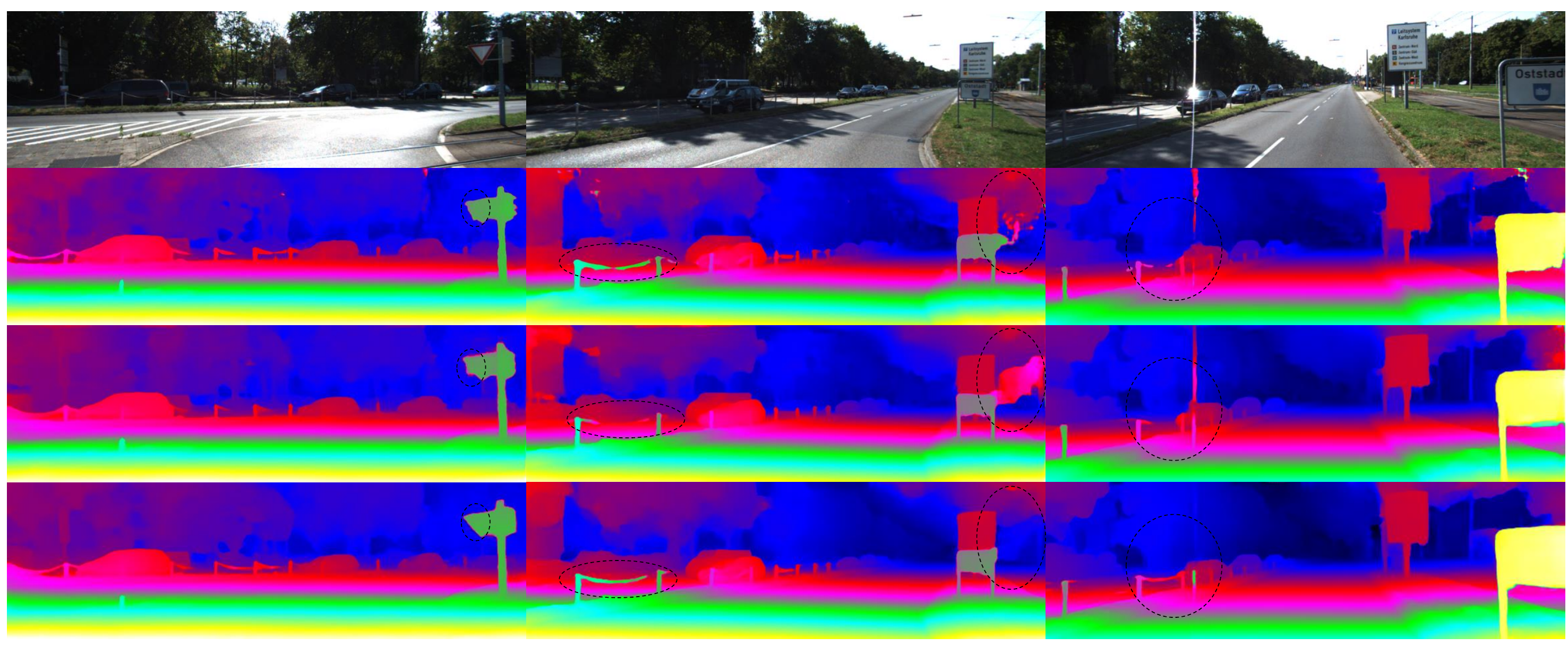}
\end{center}
\caption{Qualitative comparisons on KITTI 2015. The first row shows the input RGB images, the second row shows the results of PSMNet, the third shows the results of DeepPruner-Fast, and the last row is the results of BGNet+.}
\label{fig:kitti}
\end{figure}

In addition, we build another model variant BGNet+. Compared with BGNet, BGNet+ has an additional hourglass disparity refinement module as in \cite{AANet2020}.
The resulting model is still real-time (\textbf{30} fps). Although the model is light-weight, it even outperforms some complex networks, such as GCNet~\cite{gcnet2017}, iResNet~\cite{iresnet2018stereo} and PSMNet \cite{PSMNet2018} on the KITTI 2015 dataset. Qualitative results of BGNet+ on KITTI 2015 are shown in Figure \ref{fig:kitti}.

\begin{table*}
    \centering
    \begin{tabular}{|c|cccccc|ccc|c|}
    \hline
     & \multicolumn{6}{c|}{KITTI 2012 \cite{kitti2012}} & \multicolumn{3}{c|}{ KITTI 2015 \cite{kitti2015} } & \\
    \hline
    Method& 2-noc & 2-all & 3-noc & 3-all & \thead{EPE \\ noc} & \thead{EPE\\all} & D1-bg & D1-fg & D1-all & \makecell{Runtime \\ (ms)} \\
    \hline
    {MC-CNN-acrt \cite{mc_cnnn2016stereo}}  & 3.90 & 5.45 & 2.09 & 3.22 & 0.6  & 0.7 & 2.89 & 8.88 & 3.89 & 67000\\
    {SGM-Net \cite{sgm_net2017}}  & 3.60 & 5.15 & 2.29 & 3.50 & 0.7  & 0.9 & 2.66 & 8.64 & 3.66  & 67000\\ 
    GANet \cite{ganet2019} & \textbf{1.89} & \textbf{2.50} & \textbf{1.19} & \textbf{1.6} & 0.4 &0.5 & \textbf{1.48} & 3.46 & \textbf{1.81} & 1800 \\
    PSMNet \cite{PSMNet2018} & 2.44 & 3.01 & 1.49 & 1.89 & 0.5 &0.6 & 1.86 & 4.62 & 2.32 & 410 \\
    {GC-Net} \cite{gcnet2017} & {2.71} & {3.46} &1.77 & 2.30 & 0.6 & 0.7 & 2.21 & 6.16 & 2.87 & 900 \\
    EdgeStereo-V2 \cite{edgestereo2020} & 2.32 & 2.88 & 1.46 & 1.83 & 0.4 & 0.5  & 1.84 & \textbf{3.30} & 2.08 & 320 \\
    {iResNet2-i2 \cite{iresnet2018stereo}} & 2.69 & 3.34 & 1.71 & 2.16 & 0.5 & 0.6 & 2.25 & 3.40 & 2.44 & \textbf{120}\\
    {DeepPruner-Best\cite{deeppruner2019}} & - & - & - & - & - & - & 1.87 & 3.56 & 2.15 & 182\\
    \hline
    {DeepPruner-Fast\cite{deeppruner2019}} & - & - & - & - & - & - & 2.32 & 3.91 & 2.59 & 61\\
    RTSNet \cite{lee2019real} & 3.98 & 4.61 & 2.43 & 2.90 & 0.7 & 0.7 & 2.86 & 6.19 & 3.41 & 20 \\ 
    {StereoNet \cite{stereonet2018}} & 4.91 & 6.02 & - & - & 0.8 & 0.9 & 4.30 & 7.45 & 4.83 & \textbf{15}\\
    {DispNet \cite{dispnet2016}} & 7.38 & 8.11 & 4.11 & 4.65 & 0.9 & 1.0 & 4.32 & 4.41 & 4.34 & 60\\
     {FADNet\cite{fadnet2020}} & 3.98 & 4.63 & 2.42 &	2.86 & 0.6 & 0.7 & 2.68 & \textbf{3.50} & 2.82 & 50\\
     {AANet\cite{AANet2020}} & 2.90 &	3.60  & 1.91 &	2.42 & 0.5 & 0.6 &  1.99 & 5.39 & 2.55 & 62\\
     {BGNet} &   3.13 &   3.69 &  1.77 & 2.15 & 0.6 & 0.6 & 2.07  & 4.74  & 2.51 & 25.4\\
     {BGNet+} & \textbf{2.78} & \textbf{3.35} & \textbf{1.62} & \textbf{2.03} & 0.5 & 0.6 & \textbf{1.81}  & 4.09  & \textbf{2.19} & 32.3\\
    \hline
    \end{tabular}
    \caption{Quantitative evaluation on the test sets of KITTI 2012 and KITTI 2015. For KITTI 2012, we report the percentage of pixels with errors larger than $x$ disparities in both non-occluded (x-noc) and all regions (x-all), as well as the overall EPE in both non occluded (EPE-noc) and all the pixels (EPE-all). For KITTI 2015, we report the percentage of pixels with EPEs larger than 3 pixels in background regions (D1-bg), foreground regions (D1-fg), and all (D1-all).}
\label{tab:kitti12012_2015}
\vspace{-10pt}
\end{table*}

\textbf{Runtime analysis.} We calculated the average time of each module by testing 100 pairs of stereo images on KITTI 2015, as shown in Table \ref{tab:runtime}. The proposed cost volume upsampling module is efficient and takes 4.3 ms. 

The run-time for  bilateral grid generation, guidance image calculation and upsampling with BG is 4.3 ms, 0.07 ms, and 0.74 ms, respectively. The linear cost volume upsampling (LU) takes 3.78 ms. However, when the whole network is deployed in GPU, BG upsampling is only 0.2 ms slower than linear upsampling. Note, the run-time of the whole network is not simply the sum of run-times consumed by all modules, this may be related to the parallel computing mechanism of GPU.

\begin{table}
\begin{center}
\begin{tabular}{|c|c|c|c|c|c|}
\hline
Module & Time (ms)\\
\hline
Feature extraction &  8.8\\
Cost volume building  and aggregation &  12.2\\
Bilateral grid operation & 4.3\\
Refinement (for BGNet+) & 7.0\\
\hline
Total & 32.3\\
\hline
\end{tabular}
\end{center}
\caption{Runtime analysis for each module of BGNet and BGNet+ on the KITTI 2015 dataset. }
\label{tab:runtime}
\end{table}

\subsection{Generalization Performance}
\begin{figure*}
\begin{center}
\includegraphics[width=1.0\linewidth]{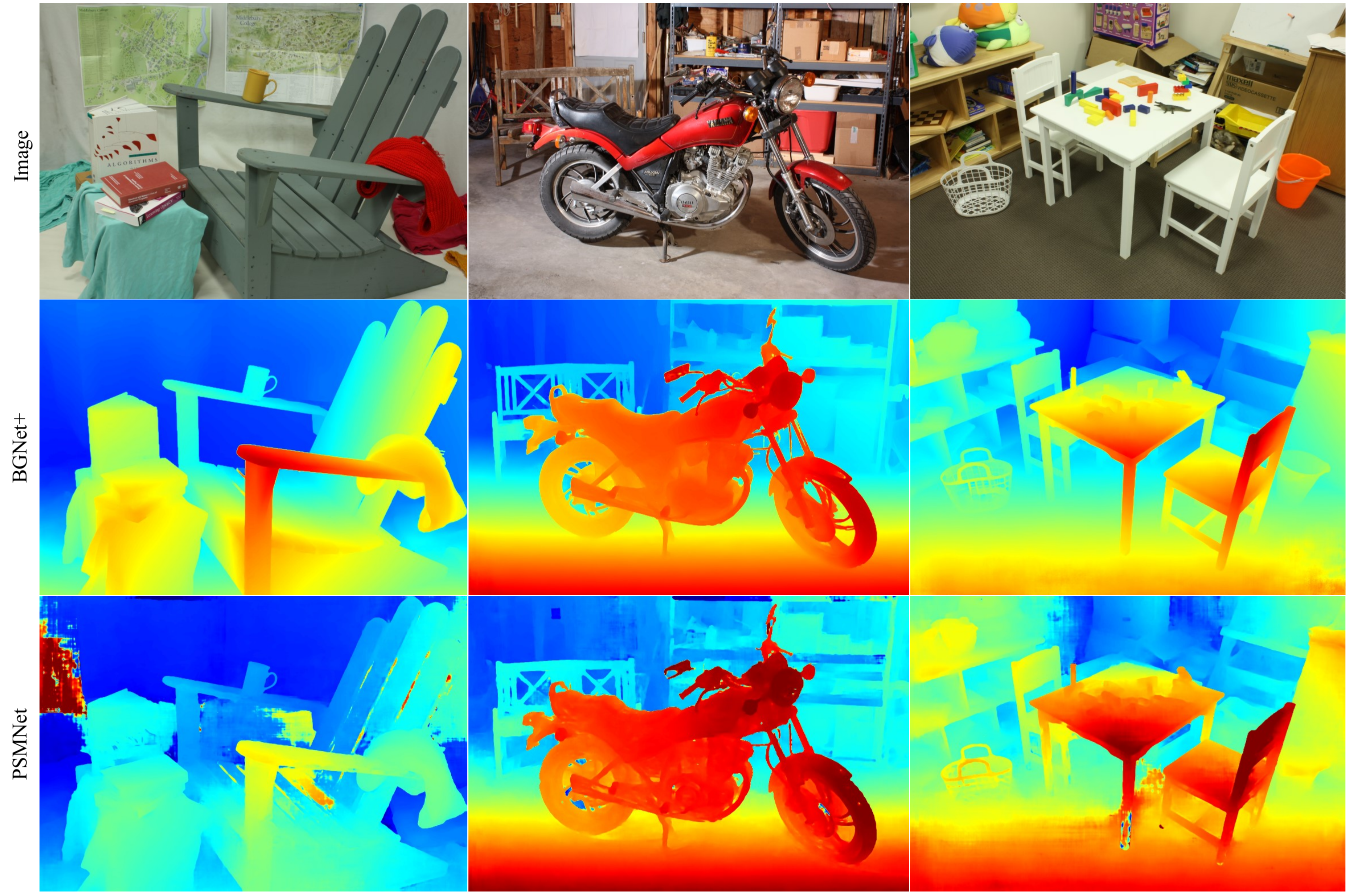}
\end{center}
\caption{Qualitative results of generalization performance evaluation on the Middlebury 2014 dataset. The first row shows the left RGB images of the stereo pairs, the second row shows the results of BGNet+, and the third row shows the results of PSMNet. The models were trained only on the synthetic SceneFlow dataset.}
\label{fig:middlebury}
\end{figure*}

The generalization performance is important for a stereo network. We evaluate the generalization ability of our network on the training set of the Middlebury 2014 dataset, where the parameters of these networks are trained on Flyingthings3D of SceneFlow \cite{dispNetC2016large} only, no additional training is done on Middlebury.
The results are illustrated in Table \ref{tab:generalization}. 
Our BGNet and BGNet+ show better performance than some complex networks, such as iResNet~\cite{iresnet2018stereo}, GANet \cite{ganet2019}, and PSMNet~\cite{PSMNet2018}. Figure \ref{fig:middlebury} shows the qualitative disparity estimation results achieved by BGNet+ and PSMNet on this benchmark. 

For DeepPruner-Fast-BG, as shown in Table \ref{tab:generalization}, the Bad 2.0 error is lower than its original version, which further demonstrates the effectiveness of our cost volume upsampling module. 

DSMNet \cite{zhang2019domain} has the best generalization performance, which benefits from feature normalization along both the spatial axis and the channel dimension. 
However, the domain invariant normalization process increases the computational time of the network. For the resolution of KITTI 2015, when the batch normalization modules are replaced by the domain invariant normalization modules, the computational time of the network is increased by 15.4 ms. 

Additionally, we conduct another interesting evaluation. We use the IRS dataset \cite{IRS2019}, a large synthetic stereo dataset, and Flyingthings3D together to train BGNet and BGNet+. IRS contains more than 100,000 pairs of $960\times540$ resolution stereo images (84,946 for training and 15,079 for testing) in indoor scenes. The Bad 2.0 error of BGNet on Middlebury reduces from \textbf{17.5} to \textbf{13.5}, and the Bad 2.0 error of BGNet+ reduces from \textbf{17.2} to \textbf{11.6}. 
This shows that the synthetic dataset also plays a significant role in improving the generalization performance of the network without increasing computational time.

\begin{table}
\begin{center}
\begin{tabular}{|c|c|c|c|c|c|}
\hline
Network & Bad 2.0 (\%)\\
\hline
PSMNet~\cite{PSMNet2018} & 25.1 \\
iRestNet-i2~\cite{iresnet2018stereo} &  19.8\\
GANet~\cite{ganet2019} & 20.3 \\
DSMNet~\cite{zhang2019domain} & \textbf{13.8} \\
DeepPruner-Fast~\cite{deeppruner2019} & 17.8\\
DeepPruner-Fast-BG & 16.5\\
BGNet & 17.5\\
BGNet+ &  17.2\\
\hline
BGNet (IRS) &  13.5\\
BGNet+ (IRS) &  \textbf{11.6}\\
\hline
\end{tabular}
\end{center}
\caption{Generalization ability on the Middlebury 2014 dataset of half resolution. The percentage of pixels with errors larger than 2 pixels (Bad 2.0) is reported. All the models are trained on the synthetic datasets.}
\label{tab:generalization}
\end{table}


\section{Conclusion}
In this paper, we have proposed a cost volume upsampling module in the learned bilateral grid. The upsampling module can be seamlessly embedded into many existing end-to-end network architectures and can  accelerate these networks significantly while maintaining comparable accuracy. In addition, based on this upsampling module, we design two real-time networks, BGNet and BGNet+. They outperform all the published networks with computational time less than 50 ms on the KITTI 2012 and KITTI 2015 datasets. The experimental results also demonstrate the good generalization ability of our network.

In the future, we plan to apply our approach to both monocular depth estimation and depth completion tasks.

\textbf{Acknowledgements.} This work was  supported by Orbbec Inc. (No. W2020JSKF0547), the National Natural Science Foundation of China (No. U20A20185, 61972435, 62076086), the Natural Science Foundation of Guangdong Province (2019A1515011271), and the Shenzhen Science and Technology Program (No. RCYX20200714114641140, JCYJ20190807152209394).

{\small
\bibliographystyle{ieee_fullname}
\bibliography{bgnet}
}

\end{document}